%% file: arxiv.tex
\documentclass{article} 
\usepackage{iclr2026_conference,times}
\usepackage{enumitem}
\usepackage{caption}
\usepackage{fancybox}

\usepackage[utf8]{inputenc} 
\usepackage[T1]{fontenc}    
\usepackage{hyperref}       
\usepackage{url}            
\usepackage{booktabs}       
\usepackage{amsfonts}       
\usepackage{nicefrac}       
\usepackage{microtype}      
\usepackage{xcolor}         
\definecolor{citecolor}{HTML}{0071BC}
\definecolor{linkcolor}{HTML}{D32F2F}
\definecolor{cellcolor}{HTML}{E3F2FD}
\definecolor{red}{HTML}{D32F2F}
\definecolor{magenta}{HTML}{D81B60}

\usepackage{booktabs}
\usepackage[table]{xcolor} 

\usepackage{amsmath}
\usepackage{amssymb}
\usepackage{mathtools}
\usepackage{amsthm}
\usepackage{siunitx} 
\usepackage[capitalize,noabbrev]{cleveref}

\theoremstyle{plain}

\theoremstyle{definition}

\theoremstyle{remark}

\usepackage[textsize=tiny]{todonotes}
\usepackage{microtype}
\usepackage{graphicx}
\usepackage{subfigure}
\usepackage{booktabs} 

\usepackage{colortbl}  
\definecolor{highlight}{RGB}{230,240,255}

\usepackage{booktabs}                    
\usepackage[table]{xcolor}               
\usepackage{array}                       
\usepackage{tabularx}                    
\usepackage{caption}                     
\usepackage{booktabs}
\usepackage{array}
\usepackage{fancybox}
\usepackage{amsmath}
\usepackage{pgfplots}
\pgfplotsset{compat = newest}
\usepackage{multirow}

\usepackage{framed}
\usepackage{tcolorbox}

\usepackage{algorithm}
\usepackage{mathrsfs}
\usepackage{mathtools}
\usepackage{algorithmic}
\usepackage{listings}
\usepackage{makecell}
\usepackage{colortbl}
\usepackage{wrapfig}
\usepackage{color}
\usepackage{wrapfig}
\usepackage{cancel}
\usepackage{soul,xcolor}
\usepackage[table]{xcolor}
\definecolor{citecolor}{HTML}{0071BC}
\definecolor{linkcolor}{HTML}{D32F2F}
\definecolor{cellcolor}{HTML}{E3F2FD}
\definecolor{blue}{HTML}{0071BC}
\definecolor{red}{HTML}{D32F2F}
\definecolor{magenta}{HTML}{D81B60}
\hypersetup{colorlinks=true, linkcolor=linkcolor, citecolor=citecolor,urlcolor=magenta}
\usepackage{pifont}
\usepackage{tikz}
\usetikzlibrary{shapes, positioning, arrows.meta, calc, decorations.pathmorphing, quotes}
\usepackage{todonotes}
\renewcommand{\cite}{\citep}
\input{math_commands.tex}

\usepackage{hyperref}
\usepackage{url}

\title{	
Revisiting Visual Understanding in Multimodal Reasoning through a Lens of Image Perturbation}
\author{%
    \textbf{Yuting Li}$^{1}$
    \quad
    \textbf{Lai Wei}$^{1,3}$
    \quad
    \textbf{Kaipeng Zheng}$^{1,2}$
    \quad
    \textbf{Jingyuan Huang}$^{1,5}$
    \quad
    \textbf{Guilin Li}$^{4}$
    \\[0.2cm]
    \textbf{Bo Wang}$^{4}$
    \quad
    \textbf{Linghe Kong}$^{1}$
    \quad
    \textbf{Lichao Sun}$^{6}$
    \quad
    \textbf{Weiran Huang}$^{1,2,5,}$\thanks{Correspondence to Weiran Huang (weiran.huang@outlook.com).}\\[0.3cm]
    $^1$ School of Computer Science, Shanghai Jiao Tong University\\[0.1cm]
    $^2$ Shanghai Innovation Institute \quad 
    $^3$ Zhongguancun Academy  \quad
    $^4$ Tencent \\[0.1cm]
    $^5$ State Key Laboratory of General Artificial Intelligence, BIGAI  \quad
    $^6$ Lehigh University
}

\iclrfinalcopy 

\begin{document}

\maketitle

\begin{abstract}
Despite the rapid progress of multimodal large language models (MLLMs), the role of visual processing in multimodal reasoning remains underexplored. In a simple yet revealing experiment, we find that language-only models, when augmented with image captions, can sometimes outperform multimodal counterparts consuming raw visual inputs. This indicates that current MLLMs may perceive visual content but fail to effectively integrate it during reasoning. Moreover, even minimal visual perturbations such as small rotations lead to severe performance drops, exposing a fragility in their visual understanding. To address this overlooked bottleneck, we propose a lightweight visual perturbation (VP) framework that strengthens perceptual robustness without architectural changes or additional data. VP introduces three targeted strategies—distractor concatenation, dominance-preserving mixup, and random rotation—that can be seamlessly integrated into post-training pipelines including SFT, DPO, and GRPO. Extensive experiments across four multimodal reasoning benchmarks show consistent absolute gains of 1–2 points, with improvements holding across datasets, training pipelines, and even advanced RL-tuned models. Ablation and task-level analyses further reveal how different perturbations uniquely benefit geometry, algebra, OCR, and chart reasoning. These findings underscore a central insight: better reasoning begins with better seeing. Our code is available at \url{https://github.com/YutingLi0606/Vision-Matters}.
\end{abstract}

\section{Introduction}
Recent advances in multimodal large language models (MLLMs) have led to impressive capabilities in vision-language understanding and reasoning ~\cite{liu2023llava, zhu2023minigpt, li2023blip, wei2023instructiongpt, wang2024qwen2vl}. Yet, their performance on math-centric reasoning tasks remains unsatisfactory, especially when visual information such as diagrams, charts, or spatial layouts is essential for problem-solving. Prior efforts have largely focused on two directions: (i) synthesizing large-scale multimodal datasets tailored for reasoning ~\cite{gao2023g, zhang2024mavis, dong2024insight}, and (ii) advancing model architectures or training objectives~\cite{meng2025mmeureka, peng2025lmmr1, deng2025openvlthinker, wei2025advancing, wei2025unsupervised}. However, a fundamental question has received little attention: how effectively do MLLMs process and integrate visual inputs during reasoning?

\begin{figure}[t]
    \centering
    \vspace{-4mm}
    \includegraphics[width=1\textwidth, trim=0cm 0cm 0cm 0cm]{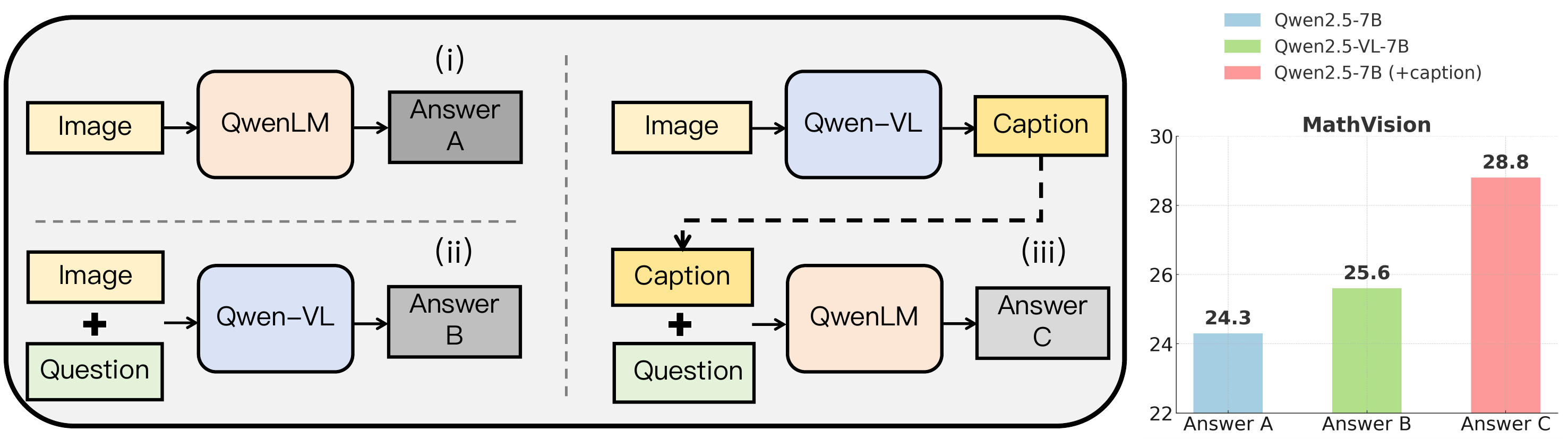}
    \caption{The left panel shows three settings: (i) Answer A (LLM-only), where a language model answers from text only; (ii) Answer B (MLLM), where a multimodal model jointly encodes both the question and the image; and (iii) Answer C (Caption-augmented LLM), where an image caption generated by the same MLLM is appended to the question for the LLM. The right panel presents quantitative results on MathVision~\cite{wang2025mathvision}. We interestingly find that language-only models, when provided with image captions, can sometimes achieve even better performance than MLLMs that consume raw visual inputs. \textbf{This suggests that current MLLMs may generate accurate visual descriptions but fail to effectively integrate them during reasoning.}}
    \label{fig:teaser}
\end{figure}

We begin with a simple yet striking observation. As illustrated in Figure~\ref{fig:teaser}, caption-augmented language models, in which an LLM is provided with captions generated by the same MLLM, can sometimes achieve comparable or even higher accuracy than the multimodal model itself. For example, on MathVision, Qwen2.5-7B with captions attains 28.8\%, surpassing Qwen2.5-VL-7B at 25.6\%. This suggests that while MLLMs can generate accurate visual descriptions, they often fail to leverage them for downstream reasoning. A second observation reinforces this diagnosis: applying benign perturbations such as random rotations, which preserve semantic content, causes large accuracy drops across multiple benchmarks; on MathVista, performance drops by 17.1 percentage points. Together, these results reveal a critical bottleneck. MLLMs are capable of perceiving images but do not robustly reason with visual information.

Motivated by this gap, we propose a lightweight visual perturbation (VP) framework that enhances perceptual robustness without introducing new data or modifying architectures. VP applies three targeted perturbations: distractor concatenation, dominance-preserving mixup, and random rotation. These are designed to challenge models’ ability to localize, filter, and reason over relevant visual features under structured variation. Crucially, VP is pipeline-agnostic and can be incorporated into existing alignment methods such as SFT~\cite{tong2024dart}, DPO~\cite{rafailov2024direct}, and GRPO~\cite{guo2025deepseekr1}.

We conduct comprehensive experiments across four datasets and consistently observe absolute performance gains of 1–2 points on average across four benchmarks (MathVision, MathVista, MathVerse, and WeMath), demonstrating that our method yields robust improvements under diverse settings. It is worth noting that, unlike recent RL-tuned models that rely on collecting additional large-scale datasets to maximize performance, we deliberately restrict training to publicly available data to provide a more rigorous and controlled validation of our approach. Beyond dataset diversity, we also verify the effectiveness of our method across different alignment pipelines, including SFT, DPO, and GRPO, showing that the improvements hold regardless of the training pipeline. Importantly, our framework also gains improvement build on current advanced models. Moreover, on the Geometry3K~\cite{lu2021inter}dataset, adding our visual perturbation strategies to Qwen2.5-VL-7B with GRPO training allows it to achieve performance comparable to state-of-the-art RL-tuned models, despite relying on far less data.

Through comprehensive ablation studies, we further highlight that not all visual perturbations are equally beneficial. Information-preserving strategies such as distractor concatenation, dominance-preserving mixup, and rotation generally improve performance, while information-degrading perturbations like Gaussian blur (--7.8\%) and random cropping 45\%(--11.5\%) cause severe declines. To further understand why this occurred, we designed a follow-up analysis breaking down performance by problem type: geometry, algebra, table \& chart, and OCR. A breakdown by problem type further shows that perturbations have task-specific effects. Rotation strongly benefits geometry reasoning (+6.8\%) but reduces accuracy on algebra, table, and OCR tasks where text readability and spatial order are crucial. In contrast, distractor concatenation and dominance-preserving mixup provide more balanced gains, consistently improving algebra and OCR while maintaining competitive performance on geometry. 
These findings highlight that the effectiveness of perturbations depends on aligning them with the visual requirements of different reasoning tasks. 
These findings demonstrate that visual perturbation plays a critical role in multimodal mathematical reasoning---better reasoning fundamentally depends on better visual understanding.

The key contributions of this work are as follows:
\begin{itemize}[itemsep=0pt,topsep=0pt,leftmargin=0.5cm]
    \item 
    We identify a critical insight: caption-augmented LLMs can match or even surpass MLLMs, while minor image perturbations lead to significant accuracy declines, underscoring fundamental weaknesses in how current MLLMs process and integrate visual information.

    \item 
    We propose a simple yet effective visual perturbation framework that consistently improves performance across datasets, pipelines, and model scales without requiring extra data;

    \item 
    We provide detailed empirical and task-level analyses, showing how different perturbations complement each other and revealing the underexplored role of visual processing in multimodal reasoning. 
\end{itemize}

\section{Related Work}
\paragraph{Multimodal Mathematical Reasoning.}
The mathematical reasoning abilities of multimodal large language models (MLLMs) have become a central research focus~\cite{zhuang2024math, gao2023text, li2024llavaonevision, dong2024insight, hu2024visual, yang2024mathglm, han24infimm, guo2024mammoth}. Compared to text-only reasoning~\cite{luo2023wizardmath, yu2023metamath}, multimodal approaches must also process visual inputs, which makes tasks such as geometry and chart interpretation substantially more challenging~\cite{chen2021geoqa}. To address these challenges, prior work has mainly advanced along two directions. First, large-scale data synthesis and task-specific dataset construction have been widely explored, e.g., MAVIS for math-centric visual data generation~\cite{zhang2024mavis}, Math-LLaVA with MathV360K~\cite{shi2024math}, Multimath with textbook data and GPT-4 validation~\cite{peng2024multimath}, and reasoning-focused datasets such as LLaVA-CoT-100k~\cite{LLaVA-CoT-abs-2411-10440} and Mulberry-260k~\cite{yao2024mulberry}. Second, architectural or algorithmic innovations have been proposed, including specialized encoders~\cite{chen2024far} and structured representations like R1-onevision~\cite{yang2025r1}. While these synthesis-driven and architecture-driven practices have led to significant progress, they rarely isolate and examine the role of visual processing itself. In contrast, our work highlights this visual bottleneck as an underexplored but critical perspective for advancing multimodal mathematical reasoning.

\paragraph{Multimodal Data Augmentation.}
Data augmentation is a common strategy to improve multimodal models. MixGen~\cite{hao2023mixgen} generates new image–text pairs by interpolating images and concatenating texts, while RobustMixGen~\cite{kim2025robustmixgen} mitigates the spurious correlations in MixGen to enhance OOD robustness. Other approaches move beyond simple input mixing: XTRA~\cite{gur2021cross} enriches training with retrieved image–caption pairs, and LEMDA~\cite{liu2022learning} learns feature-level multimodal augmentations applicable across modalities. While our work aligns with the general philosophy of these augmentation methods, existing approaches do not specifically target the emerging class of multimodal reasoning tasks for MLLMs. In contrast, we design tailored visual perturbation strategies explicitly for such reasoning tasks and demonstrate their effectiveness when integrated with various MLLM training pipelines, including SFT~\cite{tong2024dart}, DPO~\cite{rafailov2024direct}, and GRPO~\cite{guo2025deepseek}. Recently, concurrent work Noisyrollout~\cite{liu2025noisyrollout} propose a simple yet effective data augmentation method that mixes trajectories from both clean and moderately distorted images during RL training.

\section{Observation}\label{sec:observation}
\begin{table}[t]
\caption{Performance of QwenLMs and Qwen-VLs on MathVision, MathVista, MathVerse, and We-Math benchmarks. 
Star symbol ($^*$) denotes that LLMs are prompted with image captions generated by the same Qwen-VL for each question.}
\label{tab:observation}
\centering
\renewcommand{\arraystretch}{1.3}
\resizebox{0.9\textwidth}{!}{%
\begin{tabular}{l|c|cccc}
\toprule
\textbf{Models} & \textbf{Size} & \textbf{MathVision} & \textbf{MathVista} & \textbf{MathVerse} & \textbf{We-Math} \\
\midrule 
Qwen2.5-7B & 7B & 24.3 & 32.0 & 28.5 & 38.1 \\
Qwen2.5-VL-7B & 7B & 25.6 & 66.2 & 44.3 & 62.9 \\
Qwen2.5-7B$^*$ & 7B & \textbf{28.8} & 56.7 & 41.5 & 57.3 \\
\midrule
QwQ-Preview & 32B & 37.3 & 34.5 & 34.1 & 41.8 \\
QvQ-Preview & 72B & 35.6 & 71.2 & 53.2 & 68.7 \\
QwQ-Preview$^*$ & 32B & \textbf{42.9} & 63.6 & \textbf{54.9} & 61.5 \\
\midrule
\multicolumn{6}{c}{\textbf{\textit{Benchmark w/ Random Rotation} }} \\
\midrule
Qwen2.5-VL-7B & 7B & \,\,\,\,\,\,\,\,\,\,\,\,\,\,\,\,22.9 \textbf{(\textcolor{red!120}{--2.7})} & \,\,\,\,\,\,\,\,\,\,\,\,\,\,\,\,49.1 (\textcolor{red!120}{\textbf{--16.3}}) & \,\,\,\,\,\,\,\,\,\,\,\,\,\,\,\,37.5 (\textcolor{red!120}{\textbf{--6.8}}) & \,\,\,\,\,\,\,\,\,\,\,\,\,\,\,\,57.2 (\textcolor{red!120}{\textbf{--5.7}}) \\
\bottomrule
\end{tabular}
}
\end{table}
While multimodal reasoning has recently attracted significant research attention, the role of visual processing in MLLMs remains insufficiently explored. Our work begins with several simple yet revealing observations about how current MLLMs utilize visual information in reasoning tasks.
First, we evaluate three settings as illustrated in Figure~\ref{fig:teaser}:  (i) a pure language model evaluated directly on text-only questions, (ii) its multimodal counterpart that processes raw visual inputs, and (iii) the language model augmented with image captions generated by the same multimodal model (e.g., captions for Qwen2.5-7B are generated by Qwen2.5-VL-7B, and captions for QwQ-Preview are generated by Qwen2.5-VL-72B). We observe an interesting pattern in Table~\ref{tab:observation}: pure language models, when provided with image captions, can sometimes achieve comparable or even better performance than multimodal models that process raw visual inputs. 

Specifically, On MathVision~\cite{wang2024math}, the 7B language model Qwen2.5-7B~\cite{yang2024qwen2} achieves a score of 24.3, nearly matching its multimodal counterpart Qwen2.5-VL-7B~\cite{bai2025qwen2_5_vl} at 25.6. Remarkably, when augmented with captions generated by Qwen2.5-VL-7B, Qwen2.5-7B improves to 28.8, surpassing Qwen2.5-VL-7B using raw visual input. This effect is not limited to small-scale models. QwQ-Preview (32B) achieves 37.3 on MathVision~\cite{wang2025mathvision}, but rises to 42.9 with captions, exceeding the much larger 72B multimodal QvQ-Preview, which scores only 35.6. MathVerse~\cite{zhang2024mathverse} demonstrates consistent results, reinforcing that this phenomenon is not confined to a single benchmark. Through this simple exploratory experiment, it suggests that current MLLMs might not effectively integrate their visual capabilities into reasoning tasks. We hypothesize that a caption-augmented language model establishes a natural lower bound for the performance of an ideal multimodal model on visual reasoning tasks, under the assumption that both models possess comparable language understanding capabilities. Since image captions are compressed representations of visual content, they inherently contain less information than the original images. Thus, a well-aligned and effective MLLM, which can directly access and process raw visual inputs, should in principle outperform or at least match a language model that only relies on generated captions. When this expectation is not met, it suggests that the MLLM may be underutilizing visual information or that its
vision-language alignment is suboptimal. 

To further probe the insufficient utilization of visual information in current MLLMs, we conduct robustness tests by applying controlled perturbations to the visual inputs. We deliberately begin with the simplest possible perturbation: random rotation. This transformation preserves all semantic content and should not pose difficulty for a robust MLLM. As shown in Table~\ref{tab:observation}, however, this simple change leads to severe degradation. For instance,Qwen2.5-VL-7B suffers a 16.3-point decline on MathVista, while similar drops are observed on MathVision, MathVerse and We-Math. Such consistent patterns across benchmarks indicate that the issue is systematic rather than benchmark-specific. These findings highlight that current MLLMs are fragile and sensitive to visual perturbations, reinforcing our earlier observation that they fail to effectively leverage raw visual inputs in reasoning tasks. 

Based on these observations, it naturally reminds us to revisit the role of visual processing in multimodal reasoning. Rather than treating the fragility of current MLLMs under simple perturbations as a limitation to be avoided, we instead view it as an opportunity to better understand their reliance on visual inputs. By deliberately introducing structured variations, we can not only gain deeper insights into how MLLMs respond to different visual perturbations but also identify ways to strengthen the perceptual robustness.

\section{Visual Perturbation Strategies}

Building upon our earlier observation (see Section~\ref{sec:observation}) that MLLMs often underutilize visual information in multimodal reasoning, we propose a lightweight visual perturbation framework aimed at improving perceptual robustness. Our method involves applying controlled perturbations to input images that preserve core semantics while introducing visual variations. These perturbations are designed to challenge the model's ability to localize, extract, and reason over relevant visual information in the presence of noise, ambiguity, or structural shifts. 

Importantly, we deliberately avoid complex perturbation designs, ensuring reproducibility while demonstrating that even simple perturbations can already yield significant gains.

In particular, we introduce three perturbation strategies at the image level, each targeting a different aspect of visual perception and reasoning. During training, one of the three perturbations is randomly applied to each image unless otherwise specified.

\textbf{Distractor Concatenation.} Given an input image $I$, we horizontally concatenate a randomly sampled, semantically irrelevant distractor image $I'$, forming $[I; I']$. 
This strategy challenges whether the model can localize and attend to the relevant subregion of the visual input while ignoring irrelevant content. It mimics real-world settings where important information may appear alongside clutter, noise, or unrelated visual elements. Robust models should learn to suppress spurious visual signals and focus on the region aligned with the textual question.
\begin{figure}[t]
    \centering
    \vspace{-4mm}
    \includegraphics[width=1\textwidth, trim=0cm 0cm 0cm 0cm]{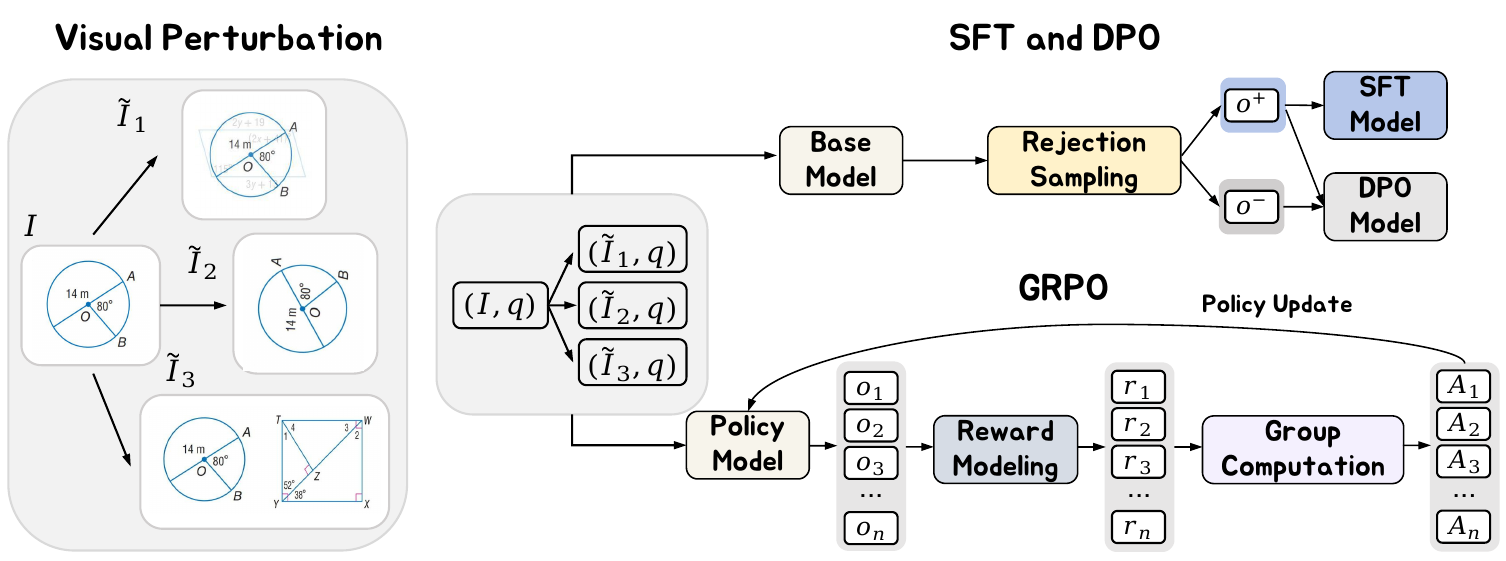}
    \caption{Our visual perturbation framework consists of three strategies: (1) distractor concatenation that horizontally combines the input image with a random distractor, (2) dominance-preserving mixup that blends the input with a distractor using skewed alpha values, and (3) random rotation that applies small angle rotations. During training, these perturbations are applied across multiple alignment pipelines including SFT, DPO, and GRPO to enhance the model's perceptual robustness and reasoning consistency.}
    \label{fig:method}
\end{figure}

\textbf{Dominance-Preserving Mixup.} Inspired by classic mixup~\cite{zhang2017mixup}, we combine the original image $I$ and a distractor $I'$ using a skewed alpha-blending: $I_{\text{mix}} = \lambda I + (1-\lambda) I'$, where $\lambda \in [0.75, 0.85]$.
Unlike standard mixup, our formulation preserves the dominant visual features of the original image while injecting low-level noise from an unrelated scene. This encourages the model to learn more invariant and robust visual features, focusing on the dominant structures relevant for  reasoning rather than overfitting to low-level image textures or noise patterns.

\textbf{Random Rotation.} 
We randomly rotate the input image to simulate geometric transformations commonly encountered in real-world diagrams and figures. This perturbation is particularly valuable for geometry-centric problems, testing the model’s spatial invariance and its ability to parse rotated structures or symbols.

\section{Experiments}
\subsection{Experimental Setup}
\noindent\textbf{Implementation Details.}
We conduct experiments using Qwen2.5-VL-7B-Instruct~\cite{bai2025qwen2_5_vl} as our base model. For SFT and DPO training, we adopt the MS-Swift~\cite{zhao2024swiftascalablelightweightinfrastructure} framework, while for GRPO training we use the EasyR1~\cite{zheng2025easyr1} framework. For SFT and DPO, we first perform rejection sampling by generating 16 responses from Qwen2.5-VL-7B-Instruct for each instruction. The responses are evaluated for correctness by comparing the extracted answers with ground truth using Qwen2.5-32B-Instruct as the evaluator. 
For SFT, we select the longest correct response as the positive sample, training for 3 epochs with a learning rate of 1e-4 and weight decay of 0.1. 
For DPO, we choose both the longest correct response as the positive sample and the shortest incorrect response as the negative sample, training for 1 epoch with a learning rate of 5e-5, weight decay of 0.1, and warmup ratio of 0.05. 
For GRPO training, we follow the default hyperparameters in EasyR1, setting training episodes to 15, using AdamW optimizer with a learning rate of \(1 \times 10^{-6}\), weight decay of \(1 \times 10^{-2}\), and gradient clipping at a maximum norm of 1.0. We set the number of rollouts per episode to 5 for GRPO training. The vision tower of Qwen2.5-VL-7B is fine-tuned without freezing, and the GRPO objective incorporates a KL divergence penalty with a coefficient of 0.01 to stabilize training. 
During training, we adopt a simple accuracy-based reward function that assigns +1 for correct final answers and 0 for incorrect ones. 
\begin{table}[t]
    \centering
    \caption{Performance comparison of GRPO training with and without visual perturbations (VP) across various training datasets. Results are evaluated on MathVision~\cite{wang2025mathvision}, MathVista~\cite{lu2023mathvista}, MathVerse~\cite{zhang2024mathverse}, and We-Math~\cite{qiao2024we} benchmarks. To make the performance gains clearer, we report the means and standard deviations over \textit{three} runs. All values represent accuracy percentages (\%). \textbf{VP consistently improves performance by 1–2 points, confirming its universality across datasets of different sizes and domains.}}
    \label{tab:main_results}
    \resizebox{\textwidth}{!}{%
    \renewcommand{\arraystretch}{1.25}
    \setlength{\tabcolsep}{6pt}
    \begin{tabular}{l l *{4}{c} | c}
        \toprule[1.2pt]
        \multirow{2}{*}{\textbf{Model and Methods}} & 
        \multirow{2}{*}{\textbf{Training Data}} &
        \multicolumn{4}{c|}{\textbf{Benchmarks}} & \multirow{2}{*}{\textbf{Average}} \\
        \cmidrule(lr){3-6}
         & & \textbf{MathVision} & \textbf{MathVista} & \textbf{MathVerse} & \textbf{We-Math} &  \\
        \midrule[0.9pt]

        \rowcolor{gray!10}
        Qwen2.5-VL-7B & -- & 25.6 & 66.2 & 44.3 & 62.9 & 49.8 \\
        \midrule[1.0pt]

        GRPO & Geometry-3K & 27.07 ± 0.38 & 69.83 ± 0.49 & 46.70 ± 0.56 & 68.63 ± 0.85 & 53.10 ± 0.40 \\
        \rowcolor{blue!5}
        \textbf{GRPO + VP} & Geometry-3K & 28.43 ± 0.75 &
72.63 ± 0.68&
48.53 ± 0.45&
70.17 ± 0.35&
\textbf{54.94 ± 0.35}
\\
        \midrule[1.0pt]
        GRPO & MMR1-6K & 29.07 ± 0.21 & 70.00 ± 0.53 & 46.03 ± 0.60 & 68.97 ± 0.67 & 53.52 ± 0.23 \\
        \rowcolor{blue!5}
        \textbf{GRPO + VP} & MMR1-6K & 31.20 ± 0.36 & 70.03 ± 0.61 & 47.63 ± 0.35 & 71.17 ± 0.42 & \textbf{55.01 ± 0.40} \\
        \midrule[1.0pt]
        GRPO & TQA-7K & 26.20 ± 0.66 & 69.17 ± 0.15 & 46.40 ± 0.26 & 66.43 ± 0.15 & 52.03 ± 0.15 \\
        \rowcolor{blue!5}
        \textbf{GRPO + VP} & TQA-7K & 26.77 ± 0.21 & 71.60 ± 0.62 & 46.43 ± 0.15 & 67.77 ± 1.19 & \textbf{53.14 ± 0.23}  \\
        \midrule[1.0pt]

        GRPO & GeoQA-8K & 27.30 ± 0.70 & 69.33 ± 0.32 & 46.93 ± 0.38 & 67.43 ± 1.27 & 52.77 ± 0.32 \\
        \rowcolor{blue!5}
        \textbf{GRPO + VP} & GeoQA-8K & 27.77 ± 0.76 & 71.97 ± 0.55 & 48.80 ± 0.56 & 70.40 ± 0.53 & \textbf{54.73 ± 0.48} \\
        
        \bottomrule[1.2pt]
    \end{tabular}
    }
\end{table}
\begin{table}[b]
    \centering
    \caption{Average accuracy (\%) of different training pipelines with and without visual perturbations (VP). For clearer presentation, results are averaged across four benchmarks. \textbf{VP consistently improves both SFT and DPO pipelines, demonstrating its pipeline-agnostic benefits.}}

    \label{tab:sft}
    \resizebox{1\textwidth}{!}{%
    \renewcommand{\arraystretch}{1}
    \setlength{\tabcolsep}{8pt}
    \begin{tabular}{l cccc c}
        \toprule[1.2pt]
        \multirow{2}{*}{\textbf{Training Method}} & 
        \multicolumn{4}{c}{\textbf{Training Datasets}} & \multirow{2}{*}{\textbf{Average}} \\
        \cmidrule(lr){2-5}
        & \textbf{Geometry-3K} & \textbf{MMR1-6K} & \textbf{TQA-7K} & \textbf{GeoQA-8K} & \\
        \midrule[0.9pt]

        
        SFT & 51.1 & 51.6 & 51.0 & 51.2   & 51.2 \\
        \rowcolor{blue!5}
        \textbf{SFT + VP} & 52.4 & 52.7 & 52.6 &  53.0  & \textbf{52.7} \\
        \midrule[0.9pt]

        
        DPO & 50.9 & 52.5 & 52.4 & 52.1 & 52.0 \\
        \rowcolor{blue!5}
        \textbf{DPO + VP} & 52.7 & 53.5 & 53.9 &  54.2   & \textbf{53.6} \\

        \bottomrule[1.2pt]
    \end{tabular}
    }
\end{table}

\noindent\textbf{Evaluation Benchmarks.}
We evaluate the MLLMs on several multimodal mathematical reasoning benchmarks: MathVision~\cite{wang2025mathvision}, MathVista~\cite{lu2023mathvista}, MathVerse~\cite{zhang2024mathverse}, We-Math~\cite{qiao2024we}. For more details about benchmarks, please
see Appendix~\ref{appendix_a}. For all benchmarks, we prompt the models to place their final answers within a designated box format. We then employ Qwen2.5-32B-Instruct~\cite{yang2024qwen2} to evaluate answer correctness by comparing the extracted responses with ground truth answers, which often contain complex mathematical expressions. Note that our reported benchmark scores may differ from those in the original papers due to variations in evaluation protocols. 

\subsection{Main Results}
\noindent\textbf{Effectiveness Across Different Training Datasets.}
Table~\ref{tab:main_results} demonstrates the effectiveness of our visual perturbation (VP) framework under GRPO training across four diverse datasets (Geometry3K~\cite{lu2021inter}, MM-R1~\cite{MMR1-Math2025}, TQA~\cite{kim2018textbook}, and GeoQA~\cite{chen2021geoqa}). To more clearly capture the performance gains introduced by VP, we report the mean and standard deviation over three independent runs.
Across all settings, incorporating VP leads to consistent accuracy gains of approximately 1–2 points over the vanilla GRPO baseline. Specifically, training on Geometry3K (3K samples) improves from 53.10\% to 54.94\%, on MM-R1 (6K samples) from 53.52\% to 55.01\%, on TQA (7K samples) from 52.03\% to 53.14\%, and on GeoQA (8K samples) from 52.77\% to 54.73\%.
These results highlight two important findings. First, the improvements are consistent across datasets of varying sizes and domains, demonstrating the universality of VP as a training enhancement. Second, the magnitude of the gains still depends on dataset quality: higher-quality datasets such as Geometry3K and GeoQA-8K show more pronounced improvements. 

\noindent\textbf{Effectiveness Across Different Training Pipelines.}
Table~\ref{tab:sft} evaluates the effectiveness of VP when applied to different alignment pipelines. We consider two commonly used training methods, supervised fine-tuning (SFT) and direct preference optimization (DPO), across four datasets. For clarity, the detailed results on each individual benchmark are provided in Appendix. The results show that VP consistently improves both pipelines. For SFT, accuracy rises from 51.2\% to 52.7\% on average, while for DPO the performance increases from 52.0\% to 53.6\%. Importantly, the gains appear across all four datasets rather than being confined to a specific training scenario, demonstrating that VP is a pipeline-agnostic enhancement. In summary, these findings confirm that VP is not tied to a particular optimization objective: whether models are trained via SFT or DPO, incorporating VP provides steady improvements and acts as a lightweight complement to existing training pipelines.

\begin{table}[t]
\centering
\captionsetup{skip=4pt} 

\begin{minipage}[t]{0.56\textwidth}
\centering
\caption{Average accuracy (\%) of different models trained with and without visual perturbations (VP). VP provides consistent gains across diverse architectures and datasets, confirming its role as a lightweight enhancement applicable to both baseline and advanced models.}
\label{tab:model_comparison}
\setlength{\tabcolsep}{2pt}
\renewcommand{\arraystretch}{1.05}
\begin{tabular*}{\textwidth}{l@{\hspace{2pt}}c@{\hspace{4pt}}c}
\toprule
\textbf{Model} & \textbf{Training Dataset} &\quad\textbf{Avg} \\
\midrule
MM-eureka-Qwen-7B     & MMK12-16K  &\quad 52.5 \\
\quad + VP            & MMK12-16K  &\quad \textbf{54.3}  \\
Qwen2.5-VL-7B         & Geometry-3K &\quad 53.1 \\
\rowcolor{blue!5}\quad + VP            & Geometry-3K &\quad\textbf{54.9} \\
ThinkLite-VL-7B       & ThinkLite-hard-11K &\quad 54.2 \\
\quad + VP            & ThinkLite-hard-11K &\quad \textbf{55.5}  \\
VL-Rethinker-7B          & ViRL-39K &\quad 55.2 \\
\quad + VP            & ViRL-39K &\quad \textbf{56.0}  \\
\bottomrule
\end{tabular*}
\end{minipage}
\hfill
\begin{minipage}[t]{0.4\textwidth}
\centering
\caption{Effect of different training data compositions with visual perturbations (VP). While moderate augmentation (e.g., Clean + 1× VP) yields the best improvements, excessive augmentation introduces redundancy and does not provide further gains.}
\label{tab:data_composition}
\setlength{\tabcolsep}{2pt}
\renewcommand{\arraystretch}{1.1}
\begin{tabular*}{\textwidth}{l@{\hspace{2pt}}c@{\hspace{4pt}}c}
\toprule
\textbf{Training Mix} & \textbf{Size} &\quad\textbf{Avg} \\
\midrule
All Clean             & 2.1k  &\quad53.1 \\
All VP                & 2.1k  &\quad54.3  \\
Half Clean + Half VP  & 2.1k  &\quad54.5  \\
\rowcolor{blue!5}Clean + 1x VP         & 4.2k  &\quad\textbf{54.9} \\
Clean + 4x VP         & 10.5k &\quad54.7 \\
\bottomrule
\end{tabular*}
\end{minipage}
\vspace{-5mm}
\end{table}

\noindent\textbf{Complementary to Advanced Models.}
We evaluate whether visual perturbations (VP) complement advanced models by continuing GRPO training from their released checkpoints on the same datasets, without introducing any new data. Concretely, MM-eureka-Qwen-7B~\cite{meng2025mmeureka}, obtained by training Qwen2.5-VL-7B on the curated MMK12-16K~\cite{meng2025mmeureka} dataset, improves from 52.5\% to 54.3\% when further trained with VP. The same holds for ThinkLite-VL-7B~\cite{wang2025sota} (trained on ThinkLite-hard-11K~\cite{wang2025sota}), which rises from 54.2\% to 55.5\%, and for VL-Rethinker-7B~\cite{wang2025vl} (trained on ViRL-39K~\cite{wang2025vl}), which increases from 55.2\% to 56.0\%. Even the base Qwen2.5-VL-7B, trained only on Geometry-3K, benefits from VP, reaching 54.9\%. These results confirm that VP acts as a lightweight and consistent enhancement, reliably adding 1–2 points on top of already strong models.

It is worth noting that the above advanced methods reach their high performance only through large-scale data collection and sophisticated algorithmic designs. In sharp contrast, our approach uses only the publicly available Geometry-3K dataset ~\cite{lu2021inter}(2.1K samples): by simply adding VP, Qwen2.5-VL-7B attains 54.9\% accuracy, which is on par with these advanced models trained on much larger and carefully engineered corpora. This demonstrates that VP not only complements existing state-of-the-art pipelines but also serves as a universal and lightweight enhancer that narrows the gap between small-data baselines and advanced systems.

\subsection{Ablation Studies}
\noindent\textbf{Different Data Compositions Using VP.}
Table~\ref{tab:data_composition} analyzes the effect of varying the proportion of visual perturbation (VP) data during training. We find that simply replacing all clean data with VP yields moderate improvement (53.1\% $\to$ 54.3\%). A balanced mix of half clean and half VP achieves slightly higher gains (54.5\%). The best result (54.9\%) comes from combining the clean set with one additional VP-augmented version (i.e., doubling the data size). Interestingly, further increasing the number of VP variants (e.g., Clean + 4x VP, totaling 10.5k samples) does not lead to additional benefits and even shows marginal decline compared to the 1x setting. These results suggest that VP is most effective when used in moderation: a small amount of augmentation is sufficient to unlock its complementary benefits, while excessive perturbation introduces redundancy without yielding further gains.

\noindent\textbf{Impact of Different Single Visual Perturbation.}
We conduct a comprehensive ablation study on different perturbation strategies to evaluate their effect on mathematical reasoning performance with GRPO training on GEOQA~\cite{chen2021geoqa}. 
Following our earlier discussion, we categorize perturbations into two groups: (1) information-preserving perturbations, which maintain the core visual semantics while introducing controlled variations, and (2) information-degrading perturbations, which deliberately distort or remove visual details. From Table~\ref{tab:grpo_ablation}, we observe a clear dichotomy. Information-degrading perturbations such as Gaussian blur, aggressive random cropping (45\%), and high-variance Gaussian noise lead to substantial performance drops, with Gaussian blur causing the largest decrease (53.1 $\rightarrow$ 49.0, --7.8\%). 
This suggests that mathematical reasoning tasks are highly sensitive to the removal of fine-grained visual details, particularly when diagrams, tables, or numbers are obscured. In contrast, our proposed information-preserving perturbations consistently improve or at least maintain performance. Overall, these findings highlight two key insights: (1) perturbations that remove or distort essential visual information harm reasoning performance, while (2) perturbations that preserve semantics but introduce controlled variation can enhance robustness and generalization. This validates the design of our three strategies as complementary approaches for strengthening the perceptual grounding of MLLMs in mathematical reasoning tasks.

\begin{figure}[t]
    \centering
    \vspace{-4mm}
    \includegraphics[width=1\textwidth, trim=0cm 0cm 0cm 0cm]{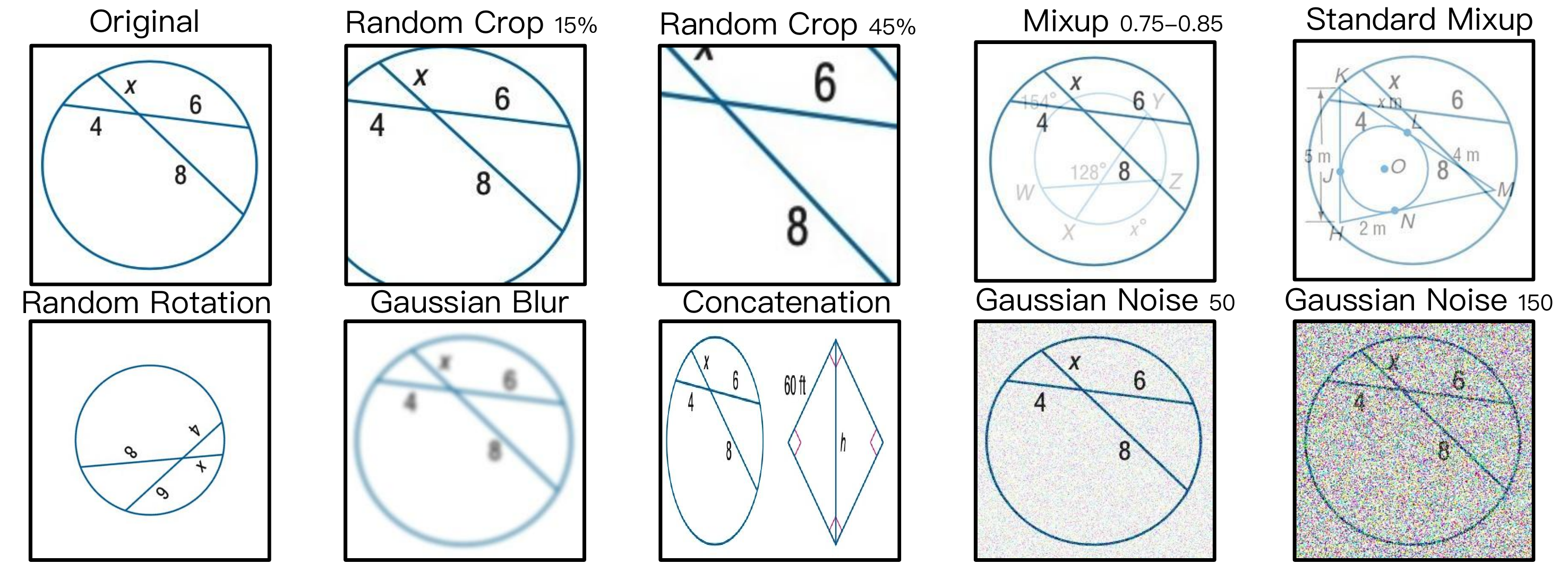}
\caption{
Visualization of different perturbation strategies used in our ablation studies. Specifically, Standard Mixup (0.45--0.55), blending two images with nearly equal weights;  Gaussian Blur, implemented with a medium kernel radius (2.5--7.5) that degrades fine details while maintaining visibility;   
Gaussian Noise, adding pixel-level noise with standard deviation 50 or 150, which disrupts low-level visual signals.  
}
    \label{fig:method}
\end{figure}

\begin{table}[t]
  \centering
  \renewcommand{\arraystretch}{1.15}
  \setlength{\tabcolsep}{3pt} 
  \caption{Performance comparison of different perturbation strategies with GRPO training across mathematical benchmarks. 
  All models are trained on the Geometry3K~\cite{chen2021geoqa} dataset. Each cell shows accuracy on the benchmark, and the average column reports overall mean and relative change.}
  \label{tab:grpo_ablation}
  \resizebox{\textwidth}{!}{%
  \begin{tabular}{>{\raggedright}p{4.4cm}@{\hspace{0.1em}} 
                      *{4}{>{\centering}p{1.7cm}} |
                      >{\centering\arraybackslash}p{2.3cm}}
  \toprule
  \multirow{2}{*}{\textbf{Perturbation Type}} & 
  \multicolumn{4}{c|}{\textbf{Benchmarks}} & \multirow{2}{*}{\textbf{Average}} \\
  \cmidrule(lr){2-5}
   & \textbf{MathVision} & \textbf{MathVista} & \textbf{MathVerse} & \textbf{WeMath} &  \\
  \midrule\rowcolor{gray!10}
  None (Baseline) & 27.1 & 69.8 & 46.7 & 68.6 & 53.1  \\
  Gaussian Blur & 24.9 & 69.5 & 40.7 & 60.8 & 49.0\,(\textcolor{red!120}{\textbf{--7.8\%}}) \\
  Random Crop 15\% & 25.4 & 70.6 & 39.8 & 65.8 & 50.4\,(\textcolor{red!120}{\textbf{--5.1\%}}) \\
  Random Crop 45\% & 22.6 & 67.1 & 37.5 & 60.9 & \,\,\,47.0\,(\textcolor{red!120}{\textbf{--11.5\%}}) \\
  Gaussian noise (std=50) & 27.1 & 69.3 & 44.1 & 67.2 & 51.9\,(\textcolor{red!120}{\textbf{--2.3\%}}) \\
  Gaussian noise (std=150) & 24.6 & 68.3 & 42.6 & 64.7 & 50.1\,(\textcolor{red!120}{\textbf{--5.7\%}}) \\
  Standard Mixup & 25.6 & 68.5 & 44.7 & 66.1 & 51.2\,(\textcolor{red!120}{\textbf{--3.6\%}}) \\
  \rowcolor{blue!5}Dominance-Preserving Mixup & 27.2 & 72.2 & 47.0 & 68.9 & 53.8\,(\textcolor{green!50!black}{\textbf{+1.3\%}}) \\
  \rowcolor{blue!5}Distractor Concatenation & 28.4 & 70.6 & 47.7 & 70.3 & 54.3\,(\textcolor{green!50!black}{\textbf{+2.3\%}}) \\
  \rowcolor{blue!5}Random Rotation & 28.0 & 71.2 & 47.5 & 69.8 & 54.1\,(\textcolor{green!50!black}{\textbf{+1.9\%}}) \\
  \bottomrule
  \end{tabular}%
  }
\end{table}

\begin{table}[t]
  \centering
  \setlength{\tabcolsep}{4pt} 
  \renewcommand{\arraystretch}{1.25}
  \small
\caption{Impact of different perturbation strategies across problem types aggregated from all benchmarks. Each cell reports \textit{correct / total} predictions and relative percentage change. Rotation benefits geometry reasoning but harms algebra, table, and OCR tasks, highlighting task-specific sensitivity to perturbations.}
  \label{tab:case_studies}
  \resizebox{\textwidth}{!}{
  \begin{tabular}{l
                  @{\hspace{4pt}}c
                  @{\hspace{4pt}}c
                  @{\hspace{4pt}}c
                  @{\hspace{4pt}}c}
  \toprule[1.2pt]
  \textbf{Perturbation Type} & \textbf{Geometry} & \textbf{Algebra} & \textbf{Table \& Chart} & \textbf{OCR} \\
  \midrule[0.9pt]
  \rowcolor{gray!10}
  Baseline & 146/381 & 103/345 & 85/242 & 52/143 \\

  Distractor Concatenation 
      & 166/381 \,(\textcolor{green!50!black}{\textbf{+5.2\%}}) 
      & 123/345 \,(\textcolor{green!50!black}{\textbf{+5.8\%}}) 
      & 96/242 \,(\textcolor{green!50!black}{\textbf{+4.5\%}}) 
      & 61/143 \,(\textcolor{green!50!black}{\textbf{+6.3\%}}) \\

  Dominance-Preserving Mixup 
      & 161/381 \,(\textcolor{green!50!black}{\textbf{+3.9\%}}) 
      & 117/345 \,(\textcolor{green!50!black}{\textbf{+4.1\%}}) 
      & 82/242 \,(\textcolor{red!120}{\textbf{--1.2\%}}) 
      & 55/143 \,(\textcolor{green!50!black}{\textbf{+2.1\%}}) \\

  Random Rotation 
      & 172/381 \,(\textcolor{green!50!black}{\textbf{+6.8\%}}) 
      & \hphantom{0}98/345 \,(\textcolor{red!120}{\textbf{--1.4\%}}) 
      & 80/242 \,(\textcolor{red!120}{\textbf{--2.1\%}}) 
      & 50/143 \,(\textcolor{red!120}{\textbf{--1.4\%}}) \\
  \bottomrule[1.2pt]
  \end{tabular}}
\end{table}

\subsection{Qualitative Analysis}

To better understand how visual perturbations influence multimodal mathematical reasoning, we further analyze their effects across four representative problem categories—geometry, algebra, table \& chart, and OCR-related tasks. While our analysis does not cover every problem type present in the benchmarks, these categories provide a diverse view of how perturbations interact with different reasoning demands.

\textbf{Geometry.} Rotation-based perturbations prove most effective here, improving accuracy from 146/381 to 172/381 (+6.8\%). This suggests that forcing the model to reason about objects under varying orientations strengthens its spatial grounding. Distractor concatenation and dominance-preserving mixup also yield solid gains (+5.2\% and +3.9\%), showing that geometry tasks generally benefit from added visual variability.

\textbf{Algebra.} Unlike geometry, algebra tasks are harmed by random rotation (103/345 $\rightarrow$ 98/345, --1.4\%), likely because rotations distort symbolic structures such as equations. In contrast, distractor concatenation (+5.8\%) and mixup (+4.1\%) both enhance performance, indicating that algebra problems benefit more from exposure to noisy but semantically consistent visual signals.

\textbf{Table \& Chart.} Perturbations are more challenging in this category. Distractor concatenation improves accuracy from 85/242 to 96/242 (+4.5\%), but mixup (82/242, --1.2\%) and rotation (80/242, --2.1\%) both degrade performance. This highlights that visual consistency and alignment are particularly important when models must parse structured tabular layouts.

\textbf{OCR.} For OCR-style tasks, distractor concatenation again provides the largest boost (52/143 $\rightarrow$ 61/143, +6.3\%), while mixup yields a modest gain (+2.1\%). Random rotation, however, slightly reduces accuracy (50/143, --1.4\%), suggesting that text recognition remains highly sensitive to orientation changes.

\section{Discussion}

In this work, our primary aim was to highlight the often-overlooked importance of visual processing in multimodal reasoning, and secondly, to demonstrate the surprising effectiveness of visual perturbation. We believe that, just as data augmentation has become a cornerstone in traditional vision tasks, the multimodal community should treat visual processing with the same level of rigor and attention. While we do not discount the remarkable progress driven by large-scale data collection and algorithm design; we argue that lightweight approaches such as visual perturbation, essentially a low-cost yet effective enhancement, deserve to be recognized as a community consensus and widely adopted. 

Building on this work, we see substantial room for extension. The most immediate direction is to design more fine-grained perturbation strategies beyond the simple forms explored here, for example by adapting them to the characteristics of training images or dynamically aligning them with the training scheduler. In addition, visual perturbation should also be considered in conjunction with different algorithmic designs, where it may complement reinforcement learning, curriculum learning, or advanced alignment techniques.

\section*{Acknowledgement}
This project is supported by the National Natural Science Foundation of China (No.\ 62406192), Opening Project of the State Key Laboratory of General Artificial Intelligence (No.\ SKLAGI2024OP12), Tencent WeChat Rhino-Bird Focused Research Program, and Doubao LLM Fund.

\bibliography{reference}
\bibliographystyle{iclr2026_conference}

\clearpage

\appendix
\section{Appendix}
\subsection{LLM USAGE}
We used large language models (LLMs) as assistive tools in the preparation of this paper. Specifically,
LLMs were employed for language editing and improving clarity. All research ideas, methodologies,
theoretical results, and experiments were conceived and conducted by the authors. The authors take
full responsibility for the content of this paper.
\subsection{Evaluation Benchmarks}
\label{appendix_a}
We evaluate the MLLMs on several multimodal mathematical reasoning benchmarks:
\begin{itemize}[itemsep=0pt,topsep=0pt,leftmargin=0.5cm]
    \item \textbf{MathVision}~\cite{wang2025mathvision} is a challenging benchmark containing 3040 mathematical problems with visual contexts from real-world math competitions across 12 grades. It covers 16 subjects over 5 difficulty levels, including specialized topics like Analytic Geometry, Combinatorial Geometry, and Topology.
    
    \item \textbf{MathVista}~\cite{lu2023mathvista} is a comprehensive benchmark for evaluating mathematical reasoning in visual contexts. It contains 1000 questions featuring diverse problem types including geometry, charts, and tables.
    
    \item \textbf{MathVerse}~\cite{zhang2024mathverse} is an all-around visual math benchmark designed for an equitable and in-depth evaluation of MLLMs. The test set contains 3940 multi-subject math problems with diagrams from publicly available sources, focusing on Plane Geometry and Solid Geometry.
    
    \item \textbf{We-Math}~\cite{qiao2024we} meticulously collect and categorize 1740 visual math problems in the test set, spanning 67 hierarchical knowledge concepts and 5 layers of knowledge granularity. 
\end{itemize}

\end{document}

%% file: math_commands.tex
\usepackage{amsmath,amsfonts,bm}

\def\eqref#1{equation~\ref{#1}}

\def\1{\bm{1}}

\DeclareMathAlphabet{\mathsfit}{\encodingdefault}{\sfdefault}{m}{sl}
\SetMathAlphabet{\mathsfit}{bold}{\encodingdefault}{\sfdefault}{bx}{n}













%% file: arxiv.bbl
\begin{thebibliography}{53}
\providecommand{\natexlab}[1]{#1}
\providecommand{\url}[1]{\texttt{#1}}
\expandafter\ifx\csname urlstyle\endcsname\relax
  \providecommand{\doi}[1]{doi: #1}\else
  \providecommand{\doi}{doi: \begingroup \urlstyle{rm}\Url}\fi

\bibitem[Bai et~al.(2025)Bai, Chen, Liu, Wang, Ge, Song, Dang, Wang, Wang, Tang, et~al.]{bai2025qwen2_5_vl}
Shuai Bai, Keqin Chen, Xuejing Liu, Jialin Wang, Wenbin Ge, Sibo Song, Kai Dang, Peng Wang, Shijie Wang, Jun Tang, et~al.
\newblock Qwen2. 5-vl technical report.
\newblock \emph{arXiv preprint arXiv:2502.13923}, 2025.

\bibitem[Chen et~al.(2021)Chen, Tang, Qin, Liang, Liu, Xing, and Lin]{chen2021geoqa}
Jiaqi Chen, Jianheng Tang, Jinghui Qin, Xiaodan Liang, Lingbo Liu, Eric Xing, and Liang Lin.
\newblock Geoqa: A geometric question answering benchmark towards multimodal numerical reasoning.
\newblock In \emph{Findings of the Association for Computational Linguistics: ACL-IJCNLP 2021}, pp.\  513--523, 2021.

\bibitem[Chen et~al.(2024)Chen, Wang, Tian, Ye, Gao, Cui, Tong, Hu, Luo, Ma, et~al.]{chen2024far}
Zhe Chen, Weiyun Wang, Hao Tian, Shenglong Ye, Zhangwei Gao, Erfei Cui, Wenwen Tong, Kongzhi Hu, Jiapeng Luo, Zheng Ma, et~al.
\newblock How far are we to gpt-4v? closing the gap to commercial multimodal models with open-source suites.
\newblock \emph{arXiv preprint arXiv:2404.16821}, 2024.

\bibitem[Deng et~al.(2025)Deng, Bansal, Yin, Peng, Wang, and Chang]{deng2025openvlthinker}
Yihe Deng, Hritik Bansal, Fan Yin, Nanyun Peng, Wei Wang, and Kai-Wei Chang.
\newblock Openvlthinker: An early exploration to complex vision-language reasoning via iterative self-improvement.
\newblock \emph{arXiv preprint arXiv:2503.17352}, 2025.

\bibitem[Dong et~al.(2024)Dong, Liu, Sun, Yang, Hu, Rao, and Liu]{dong2024insight}
Yuhao Dong, Zuyan Liu, Hai-Long Sun, Jingkang Yang, Winston Hu, Yongming Rao, and Ziwei Liu.
\newblock Insight-v: Exploring long-chain visual reasoning with multimodal large language models.
\newblock \emph{arXiv preprint arXiv:2411.14432}, 2024.

\bibitem[Gao et~al.(2023{\natexlab{a}})Gao, Wang, Li, Sun, Qian, Ding, and Zhou]{gao2023text}
Dawei Gao, Haibin Wang, Yaliang Li, Xiuyu Sun, Yichen Qian, Bolin Ding, and Jingren Zhou.
\newblock Text-to-sql empowered by large language models: A benchmark evaluation.
\newblock \emph{arXiv preprint arXiv:2308.15363}, 2023{\natexlab{a}}.

\bibitem[Gao et~al.(2023{\natexlab{b}})Gao, Pi, Zhang, Ye, Zhong, Wang, Hong, Han, Xu, Li, et~al.]{gao2023g}
Jiahui Gao, Renjie Pi, Jipeng Zhang, Jiacheng Ye, Wanjun Zhong, Yufei Wang, Lanqing Hong, Jianhua Han, Hang Xu, Zhenguo Li, et~al.
\newblock G-llava: Solving geometric problem with multi-modal large language model.
\newblock \emph{arXiv preprint arXiv:2312.11370}, 2023{\natexlab{b}}.

\bibitem[Guo et~al.(2025{\natexlab{a}})Guo, Yang, Zhang, Song, Zhang, Xu, Zhu, Ma, Wang, Bi, et~al.]{guo2025deepseek}
Daya Guo, Dejian Yang, Haowei Zhang, Junxiao Song, Ruoyu Zhang, Runxin Xu, Qihao Zhu, Shirong Ma, Peiyi Wang, Xiao Bi, et~al.
\newblock Deepseek-r1: Incentivizing reasoning capability in llms via reinforcement learning.
\newblock \emph{arXiv preprint arXiv:2501.12948}, 2025{\natexlab{a}}.

\bibitem[Guo et~al.(2025{\natexlab{b}})Guo, Yang, Zhang, Song, Zhang, Xu, Zhu, Ma, Wang, Bi, et~al.]{guo2025deepseekr1}
Daya Guo, Dejian Yang, Haowei Zhang, Junxiao Song, Ruoyu Zhang, Runxin Xu, Qihao Zhu, Shirong Ma, Peiyi Wang, Xiao Bi, et~al.
\newblock Deepseek-r1: Incentivizing reasoning capability in llms via reinforcement learning.
\newblock \emph{arXiv preprint arXiv:2501.12948}, 2025{\natexlab{b}}.

\bibitem[Guo et~al.(2024)Guo, Zheng, Bai, Li, Wang, Zhu, Li, Neubig, Chen, and Yue]{guo2024mammoth}
Jarvis Guo, Tuney Zheng, Yuelin Bai, Bo~Li, Yubo Wang, King Zhu, Yizhi Li, Graham Neubig, Wenhu Chen, and Xiang Yue.
\newblock Mammoth-vl: Eliciting multimodal reasoning with instruction tuning at scale.
\newblock \emph{arXiv preprint arXiv:2412.05237}, 2024.

\bibitem[Gur et~al.(2021)Gur, Neverova, Stauffer, Lim, Kiela, and Reiter]{gur2021cross}
Shir Gur, Natalia Neverova, Chris Stauffer, Ser-Nam Lim, Douwe Kiela, and Austin Reiter.
\newblock Cross-modal retrieval augmentation for multi-modal classification.
\newblock \emph{arXiv preprint arXiv:2104.08108}, 2021.

\bibitem[Han et~al.(2024)Han, Jian, Hu, Liu, Wang, Fan, Ai, Huang, He, Yang, et~al.]{han24infimm}
Xiaotian Han, Yiren Jian, Xuefeng Hu, Haogeng Liu, Yiqi Wang, Qihang Fan, Yuang Ai, Huaibo Huang, Ran He, Zhenheng Yang, et~al.
\newblock Infimm-webmath-40b: Advancing multimodal pre-training for enhanced mathematical reasoning.
\newblock In \emph{The 4th Workshop on Mathematical Reasoning and AI at NeurIPS'24}, 2024.

\bibitem[Hao et~al.(2023)Hao, Zhu, Appalaraju, Zhang, Zhang, Li, and Li]{hao2023mixgen}
Xiaoshuai Hao, Yi~Zhu, Srikar Appalaraju, Aston Zhang, Wanqian Zhang, Bo~Li, and Mu~Li.
\newblock Mixgen: A new multi-modal data augmentation.
\newblock In \emph{Proceedings of the IEEE/CVF winter conference on applications of computer vision}, pp.\  379--389, 2023.

\bibitem[Hu et~al.(2024)Hu, Shi, Fu, Roth, Ostendorf, Zettlemoyer, Smith, and Krishna]{hu2024visual}
Yushi Hu, Weijia Shi, Xingyu Fu, Dan Roth, Mari Ostendorf, Luke Zettlemoyer, Noah~A Smith, and Ranjay Krishna.
\newblock Visual sketchpad: Sketching as a visual chain of thought for multimodal language models.
\newblock \emph{arXiv preprint arXiv:2406.09403}, 2024.

\bibitem[Kim et~al.(2018)Kim, Kim, and Kwak]{kim2018textbook}
Daesik Kim, Seonhoon Kim, and Nojun Kwak.
\newblock Textbook question answering with multi-modal context graph understanding and self-supervised open-set comprehension.
\newblock \emph{arXiv preprint arXiv:1811.00232}, 2018.

\bibitem[Kim et~al.(2025)Kim, Im, Lee, Lee, and Kang]{kim2025robustmixgen}
Sunwoo Kim, Hun Im, Woojun Lee, Seonggye Lee, and Pilsung Kang.
\newblock Robustmixgen: Data augmentation for enhancing robustness of visual--language models in the presence of distribution shift.
\newblock \emph{Neurocomputing}, 619:\penalty0 129167, 2025.

\bibitem[Leng(2025)]{MMR1-Math2025}
Sicong Leng.
\newblock Mmr1: Advancing the frontiers of multimodal reasoning.
\newblock \url{https://github.com/LengSicong/MMR1}, 2025.

\bibitem[Li et~al.(2024)Li, Zhang, Guo, Zhang, Li, Zhang, Zhang, Zhang, Li, Liu, et~al.]{li2024llavaonevision}
Bo~Li, Yuanhan Zhang, Dong Guo, Renrui Zhang, Feng Li, Hao Zhang, Kaichen Zhang, Peiyuan Zhang, Yanwei Li, Ziwei Liu, et~al.
\newblock Llava-onevision: Easy visual task transfer.
\newblock \emph{arXiv preprint arXiv:2408.03326}, 2024.

\bibitem[Li et~al.(2023)Li, Li, Savarese, and Hoi]{li2023blip}
Junnan Li, Dongxu Li, Silvio Savarese, and Steven Hoi.
\newblock Blip-2: Bootstrapping language-image pre-training with frozen image encoders and large language models.
\newblock In \emph{International conference on machine learning}, pp.\  19730--19742. PMLR, 2023.

\bibitem[Liu et~al.(2023)Liu, Li, Wu, and Lee]{liu2023llava}
Haotian Liu, Chunyuan Li, Qingyang Wu, and Yong~Jae Lee.
\newblock Visual instruction tuning.
\newblock \emph{arXiv preprint arXiv:2304.08485}, 2023.

\bibitem[Liu et~al.(2025)Liu, Ni, Wu, Du, Dou, Wang, Pang, and Shieh]{liu2025noisyrollout}
Xiangyan Liu, Jinjie Ni, Zijian Wu, Chao Du, Longxu Dou, Haonan Wang, Tianyu Pang, and Michael~Qizhe Shieh.
\newblock Noisyrollout: Reinforcing visual reasoning with data augmentation.
\newblock \emph{arXiv preprint arXiv:2504.13055}, 2025.

\bibitem[Liu et~al.(2022)Liu, Tang, Shi, Zhang, Li, Shrivastava, and Wilson]{liu2022learning}
Z~Liu, Z~Tang, X~Shi, A~Zhang, M~Li, A~Shrivastava, and AG~Wilson.
\newblock Learning multimodal data augmentation in feature space. arxiv, 2022.

\bibitem[Lu et~al.(2021)Lu, Gong, Jiang, Qiu, Huang, Liang, and Zhu]{lu2021inter}
Pan Lu, Ran Gong, Shibiao Jiang, Liang Qiu, Siyuan Huang, Xiaodan Liang, and Song-Chun Zhu.
\newblock Inter-gps: Interpretable geometry problem solving with formal language and symbolic reasoning.
\newblock \emph{arXiv preprint arXiv:2105.04165}, 2021.

\bibitem[Lu et~al.(2023)Lu, Bansal, Xia, Liu, Li, Hajishirzi, Cheng, Chang, Galley, and Gao]{lu2023mathvista}
Pan Lu, Hritik Bansal, Tony Xia, Jiacheng Liu, Chunyuan Li, Hannaneh Hajishirzi, Hao Cheng, Kai-Wei Chang, Michel Galley, and Jianfeng Gao.
\newblock Mathvista: Evaluating mathematical reasoning of foundation models in visual contexts.
\newblock \emph{arXiv preprint arXiv:2310.02255}, 2023.

\bibitem[Luo et~al.(2023)Luo, Sun, Xu, Zhao, Lou, Tao, Geng, Lin, Chen, and Zhang]{luo2023wizardmath}
Haipeng Luo, Qingfeng Sun, Can Xu, Pu~Zhao, Jianguang Lou, Chongyang Tao, Xiubo Geng, Qingwei Lin, Shifeng Chen, and Dongmei Zhang.
\newblock Wizardmath: Empowering mathematical reasoning for large language models via reinforced evol-instruct.
\newblock \emph{arXiv preprint arXiv:2308.09583}, 2023.

\bibitem[Meng et~al.(2025)Meng, Du, Liu, Zhou, Lu, Fu, Shi, Wang, He, Zhang, Luo, Qiao, Zhang, and Shao]{meng2025mmeureka}
Fanqing Meng, Lingxiao Du, Zongkai Liu, Zhixiang Zhou, Quanfeng Lu, Daocheng Fu, Botian Shi, Wenhai Wang, Junjun He, Kaipeng Zhang, Ping Luo, Yu~Qiao, Qiaosheng Zhang, and Wenqi Shao.
\newblock Mm-eureka: Exploring visual aha moment with rule-based large-scale reinforcement learning, 2025.
\newblock URL \url{https://github.com/ModalMinds/MM-EUREKA}.

\bibitem[Peng et~al.(2024)Peng, Fu, Gao, Zhong, Fu, and Tang]{peng2024multimath}
Shuai Peng, Di~Fu, Liangcai Gao, Xiuqin Zhong, Hongguang Fu, and Zhi Tang.
\newblock Multimath: Bridging visual and mathematical reasoning for large language models.
\newblock \emph{arXiv preprint arXiv:2409.00147}, 2024.

\bibitem[Qiao et~al.(2024)Qiao, Tan, Dong, Wu, Sun, Song, GongQue, Lei, Wei, Zhang, et~al.]{qiao2024we}
Runqi Qiao, Qiuna Tan, Guanting Dong, Minhui Wu, Chong Sun, Xiaoshuai Song, Zhuoma GongQue, Shanglin Lei, Zhe Wei, Miaoxuan Zhang, et~al.
\newblock We-math: Does your large multimodal model achieve human-like mathematical reasoning?
\newblock \emph{arXiv preprint arXiv:2407.01284}, 2024.

\bibitem[Rafailov et~al.(2024)Rafailov, Sharma, Mitchell, Manning, Ermon, and Finn]{rafailov2024direct}
Rafael Rafailov, Archit Sharma, Eric Mitchell, Christopher~D Manning, Stefano Ermon, and Chelsea Finn.
\newblock Direct preference optimization: Your language model is secretly a reward model.
\newblock \emph{Advances in Neural Information Processing Systems}, 36, 2024.

\bibitem[Shi et~al.(2024)Shi, Hu, Bin, Liu, Yang, Ng, Bing, and Lee]{shi2024math}
Wenhao Shi, Zhiqiang Hu, Yi~Bin, Junhua Liu, Yang Yang, See-Kiong Ng, Lidong Bing, and Roy Ka-Wei Lee.
\newblock Math-llava: Bootstrapping mathematical reasoning for multimodal large language models.
\newblock \emph{arXiv preprint arXiv:2406.17294}, 2024.

\bibitem[Tong et~al.(2024)Tong, Zhang, Wang, Wu, and He]{tong2024dart}
Yuxuan Tong, Xiwen Zhang, Rui Wang, Ruidong Wu, and Junxian He.
\newblock Dart-math: Difficulty-aware rejection tuning for mathematical problem-solving.
\newblock \emph{Advances in Neural Information Processing Systems}, 37:\penalty0 7821--7846, 2024.

\bibitem[Wang et~al.(2025{\natexlab{a}})Wang, Qu, Huang, Chu, Lin, and Chen]{wang2025vl}
Haozhe Wang, Chao Qu, Zuming Huang, Wei Chu, Fangzhen Lin, and Wenhu Chen.
\newblock Vl-rethinker: Incentivizing self-reflection of vision-language models with reinforcement learning.
\newblock \emph{arXiv preprint arXiv:2504.08837}, 2025{\natexlab{a}}.

\bibitem[Wang et~al.(2025{\natexlab{b}})Wang, Pan, Shi, Lu, Ren, Zhou, Zhan, and Li]{wang2025mathvision}
Ke~Wang, Junting Pan, Weikang Shi, Zimu Lu, Houxing Ren, Aojun Zhou, Mingjie Zhan, and Hongsheng Li.
\newblock Measuring multimodal mathematical reasoning with math-vision dataset.
\newblock \emph{Advances in Neural Information Processing Systems}, 37:\penalty0 95095--95169, 2025{\natexlab{b}}.

\bibitem[Wang et~al.(2024{\natexlab{a}})Wang, Li, Shao, Xu, Dai, Li, Chen, Wu, and Sui]{wang2024math}
Peiyi Wang, Lei Li, Zhihong Shao, Runxin Xu, Damai Dai, Yifei Li, Deli Chen, Yu~Wu, and Zhifang Sui.
\newblock Math-shepherd: Verify and reinforce llms step-by-step without human annotations.
\newblock In \emph{Proceedings of the 62nd Annual Meeting of the Association for Computational Linguistics (Volume 1: Long Papers)}, pp.\  9426--9439, 2024{\natexlab{a}}.

\bibitem[Wang et~al.(2024{\natexlab{b}})Wang, Bai, Tan, Wang, Fan, Bai, Chen, Liu, Wang, Ge, et~al.]{wang2024qwen2vl}
Peng Wang, Shuai Bai, Sinan Tan, Shijie Wang, Zhihao Fan, Jinze Bai, Keqin Chen, Xuejing Liu, Jialin Wang, Wenbin Ge, et~al.
\newblock Qwen2-vl: Enhancing vision-language model's perception of the world at any resolution.
\newblock \emph{arXiv preprint arXiv:2409.12191}, 2024{\natexlab{b}}.

\bibitem[Wang et~al.(2025{\natexlab{c}})Wang, Yang, Feng, Lu, Li, Lin, Lin, Huang, and Wang]{wang2025sota}
Xiyao Wang, Zhengyuan Yang, Chao Feng, Hongjin Lu, Linjie Li, Chung-Ching Lin, Kevin Lin, Furong Huang, and Lijuan Wang.
\newblock Sota with less: Mcts-guided sample selection for data-efficient visual reasoning self-improvement.
\newblock \emph{arXiv preprint arXiv:2504.07934}, 2025{\natexlab{c}}.

\bibitem[Wei et~al.(2023)Wei, Jiang, Huang, and Sun]{wei2023instructiongpt}
Lai Wei, Zihao Jiang, Weiran Huang, and Lichao Sun.
\newblock Instructiongpt-4: A 200-instruction paradigm for fine-tuning minigpt-4.
\newblock \emph{arXiv preprint arXiv:2308.12067}, 2023.

\bibitem[Wei et~al.(2025{\natexlab{a}})Wei, Li, Wang, Wang, Kong, Huang, and Sun]{wei2025unsupervised}
Lai Wei, Yuting Li, Chen Wang, Yue Wang, Linghe Kong, Weiran Huang, and Lichao Sun.
\newblock Unsupervised post-training for multi-modal llm reasoning via grpo.
\newblock \emph{arXiv preprint arXiv:2505.22453}, 2025{\natexlab{a}}.

\bibitem[Wei et~al.(2025{\natexlab{b}})Wei, Li, Zheng, Wang, Wang, Kong, Sun, and Huang]{wei2025advancing}
Lai Wei, Yuting Li, Kaipeng Zheng, Chen Wang, Yue Wang, Linghe Kong, Lichao Sun, and Weiran Huang.
\newblock Advancing multimodal reasoning via reinforcement learning with cold start.
\newblock \emph{arXiv preprint arXiv:2505.22334}, 2025{\natexlab{b}}.

\bibitem[Xu et~al.(2024)Xu, Jin, Li, Song, Sun, and Yuan]{LLaVA-CoT-abs-2411-10440}
Guowei Xu, Peng Jin, Hao Li, Yibing Song, Lichao Sun, and Li~Yuan.
\newblock Llava-cot: Let vision language models reason step-by-step.
\newblock \emph{CoRR}, abs/2411.10440, 2024.

\bibitem[Yang et~al.(2024{\natexlab{a}})Yang, Yang, Zhang, Hui, Zheng, Yu, Li, Liu, Huang, Wei, et~al.]{yang2024qwen2}
An~Yang, Baosong Yang, Beichen Zhang, Binyuan Hui, Bo~Zheng, Bowen Yu, Chengyuan Li, Dayiheng Liu, Fei Huang, Haoran Wei, et~al.
\newblock Qwen2. 5 technical report.
\newblock \emph{arXiv preprint arXiv:2412.15115}, 2024{\natexlab{a}}.

\bibitem[Yang et~al.(2025)Yang, He, Pan, Jiang, Deng, Yang, Lu, Yin, Rao, Zhu, et~al.]{yang2025r1}
Yi~Yang, Xiaoxuan He, Hongkun Pan, Xiyan Jiang, Yan Deng, Xingtao Yang, Haoyu Lu, Dacheng Yin, Fengyun Rao, Minfeng Zhu, et~al.
\newblock R1-onevision: Advancing generalized multimodal reasoning through cross-modal formalization.
\newblock \emph{arXiv preprint arXiv:2503.10615}, 2025.

\bibitem[Yang et~al.(2024{\natexlab{b}})Yang, Chen, Du, Yu, Wang, Hong, Jiang, Xu, Dong, and Tang]{yang2024mathglm}
Zhen Yang, Jinhao Chen, Zhengxiao Du, Wenmeng Yu, Weihan Wang, Wenyi Hong, Zhihuan Jiang, Bin Xu, Yuxiao Dong, and Jie Tang.
\newblock Mathglm-vision: Solving mathematical problems with multi-modal large language model.
\newblock \emph{arXiv preprint arXiv:2409.13729}, 2024{\natexlab{b}}.

\bibitem[Yao et~al.(2024)Yao, Huang, Wu, Zhang, Wang, Liu, Wang, Song, Feng, Shen, et~al.]{yao2024mulberry}
Huanjin Yao, Jiaxing Huang, Wenhao Wu, Jingyi Zhang, Yibo Wang, Shunyu Liu, Yingjie Wang, Yuxin Song, Haocheng Feng, Li~Shen, et~al.
\newblock Mulberry: Empowering mllm with o1-like reasoning and reflection via collective monte carlo tree search.
\newblock \emph{arXiv preprint arXiv:2412.18319}, 2024.

\bibitem[Yaowei et~al.(2025)Yaowei, Junting, Shenzhi, Zhangchi, Dongdong, and Yuwen]{zheng2025easyr1}
Zheng Yaowei, Lu~Junting, Wang Shenzhi, Feng Zhangchi, Kuang Dongdong, and Xiong Yuwen.
\newblock Easyr1: An efficient, scalable, multi-modality rl training framework.
\newblock \url{https://github.com/hiyouga/EasyR1}, 2025.

\bibitem[Yingzhe et~al.(2025)Yingzhe, Gongrui, Miaosen, Zhiyuan, Jie, Qipeng, Kai, Xingzhong, Xin, and Xu]{peng2025lmmr1}
Peng Yingzhe, Zhang Gongrui, Zhang Miaosen, You Zhiyuan, Liu Jie, Zhu Qipeng, Yang Kai, Xu~Xingzhong, Geng Xin, and Yang Xu.
\newblock Lmm-r1: Empowering 3b lmms with strong reasoning abilities through two-stage rule-based rl, 2025.

\bibitem[Yu et~al.(2023)Yu, Jiang, Shi, Yu, Liu, Zhang, Kwok, Li, Weller, and Liu]{yu2023metamath}
Longhui Yu, Weisen Jiang, Han Shi, Jincheng Yu, Zhengying Liu, Yu~Zhang, James~T Kwok, Zhenguo Li, Adrian Weller, and Weiyang Liu.
\newblock Metamath: Bootstrap your own mathematical questions for large language models.
\newblock \emph{arXiv preprint arXiv:2309.12284}, 2023.

\bibitem[Zhang et~al.(2017)Zhang, Cisse, Dauphin, and Lopez-Paz]{zhang2017mixup}
Hongyi Zhang, Moustapha Cisse, Yann~N Dauphin, and David Lopez-Paz.
\newblock mixup: Beyond empirical risk minimization.
\newblock \emph{arXiv preprint arXiv:1710.09412}, 2017.

\bibitem[Zhang et~al.(2024{\natexlab{a}})Zhang, Jiang, Zhang, Lin, Guo, Qiu, Zhou, Lu, Chang, Qiao, et~al.]{zhang2024mathverse}
Renrui Zhang, Dongzhi Jiang, Yichi Zhang, Haokun Lin, Ziyu Guo, Pengshuo Qiu, Aojun Zhou, Pan Lu, Kai-Wei Chang, Yu~Qiao, et~al.
\newblock Mathverse: Does your multi-modal llm truly see the diagrams in visual math problems?
\newblock In \emph{European Conference on Computer Vision}, pp.\  169--186. Springer, 2024{\natexlab{a}}.

\bibitem[Zhang et~al.(2024{\natexlab{b}})Zhang, Wei, Jiang, Guo, Li, Zhang, Tong, Liu, Zhou, Wei, et~al.]{zhang2024mavis}
Renrui Zhang, Xinyu Wei, Dongzhi Jiang, Ziyu Guo, Shicheng Li, Yichi Zhang, Chengzhuo Tong, Jiaming Liu, Aojun Zhou, Bin Wei, et~al.
\newblock Mavis: Mathematical visual instruction tuning with an automatic data engine.
\newblock \emph{arXiv preprint arXiv:2407.08739}, 2024{\natexlab{b}}.

\bibitem[Zhao et~al.(2024)Zhao, Huang, Hu, Wang, Mao, Zhang, Jiang, Wu, Ai, Wang, Zhou, and Chen]{zhao2024swiftascalablelightweightinfrastructure}
Yuze Zhao, Jintao Huang, Jinghan Hu, Xingjun Wang, Yunlin Mao, Daoze Zhang, Zeyinzi Jiang, Zhikai Wu, Baole Ai, Ang Wang, Wenmeng Zhou, and Yingda Chen.
\newblock Swift:a scalable lightweight infrastructure for fine-tuning, 2024.
\newblock URL \url{https://arxiv.org/abs/2408.05517}.

\bibitem[Zhu et~al.(2023)Zhu, Chen, Shen, Li, and Elhoseiny]{zhu2023minigpt}
Deyao Zhu, Jun Chen, Xiaoqian Shen, Xiang Li, and Mohamed Elhoseiny.
\newblock Minigpt-4: Enhancing vision-language understanding with advanced large language models.
\newblock \emph{arXiv preprint arXiv:2304.10592}, 2023.

\bibitem[Zhuang et~al.(2024)Zhuang, Huang, Zhang, and Zeng]{zhuang2024math}
Wenwen Zhuang, Xin Huang, Xiantao Zhang, and Jin Zeng.
\newblock Math-puma: Progressive upward multimodal alignment to enhance mathematical reasoning.
\newblock \emph{arXiv preprint arXiv:2408.08640}, 2024.

\end{thebibliography}
